\documentclass[a4paper,fleqn]{cas-dc}

\usepackage[authoryear,longnamesfirst]{natbib}
\usepackage[linesnumbered,ruled,vlined]{algorithm2e}

\def\tsc#1{\csdef{#1}{\textsc{\lowercase{#1}}\xspace}}
\tsc{WGM}
\tsc{QE}

\begin{document}
\let\WriteBookmarks\relax
\def\floatpagepagefraction{1}
\def\textpagefraction{.001}

\shorttitle{}    

\shortauthors{Zhang Liqiang, et~al.}  

\title [mode = title]{A Visual-inertial Localization Algorithm using Opportunistic Visual Beacons and Dead-Reckoning for GNSS-Denied Large-scale Applications}  

\tnotemark[1] 

\tnotetext[1]{} 

%

\author[1]{Liqiang Zhang}

\fnmark[1]

\ead{zhangliqiang@cczu.edu.cn}

\ead[url]{}

\credit{}

\affiliation[1]{{Wangzheng School of Microelectronics,Changzhou University},
            addressline={No. 2468, West Yanzheng Avenue}, 
            city={Changzhou},
            postcode={213159}, 
            state={Jiangsu Province},
            country={China}}

\author[2]{Ye Tian}

\cormark[1]
\fnmark[2]

\ead{tianye_ti@kmj.iis.u-tokyo.ac.jp}

\ead[url]{}

\credit{}

\affiliation[2]{organization={Institute of Industrial Science, The University of Tokyo},
            addressline={4-6-1 Komaba, Meguro-ku}, 
            city={Tokyo},
            postcode={153-8505}, 
            country={Japan}}

\cortext[1]{Corresponding author}

\fntext[2]{}

\author[3]{Dongyan Wei}

\fnmark[3]

\ead{weidy@aircas.ac.cn}

\ead[url]{}

\credit{}

\affiliation[3]{{Aerospace Information Research Institute, Chinese Academy of Sciences},
            addressline={No. 9 Dengzhuang South Road}, 
            city={Beijing},
            postcode={100094}, 
            country={China}}

\begin{abstract}
With the development of smart cities, the demand for continuous pedestrian navigation in large-scale urban environments has significantly increased. While global navigation satellite systems (GNSS) provide low-cost and reliable positioning services, they are often hindered in complex urban canyon environments. Thus, exploring opportunistic signals for positioning in urban areas has become a key solution. Augmented reality allows pedestrians to acquire real-time visual information. Accordingly, we propose a low-cost visual-inertial positioning solution. This method comprises a lightweight multi-scale group convolution (MSGC)-based visual place recognition (VPR) neural network, a magnetic disturbance rejection (MDR)-based pedestrian dead reckoning (PDR) algorithm, and a visual/inertial fusion approach based on a Kalman filter with gross error suppression. The VPR serves as a conditional observation to the Kalman filter, effectively correcting the errors accumulated through the MDR-PDR method. This enables the entire algorithm to ensure the reliability of long-term positioning in GNSS-denied areas. Extensive experimental results demonstrate that our method maintains stable positioning during large-scale movements. Compared to the lightweight MobileNetV3-based VPR method, our proposed VPR solution improves Recall@1 by at least 3\% on two public datasets while reducing the number of parameters by 63.37\%. It even achieves performance that is comparable to the larger VGG16-based method on the Pitts30k-test dataset, all with approximately 1.11 M parameters. In addition, the PDR-VPR algorithm improves 75\% localization accuracy by more than 40\% compared to the MDR-PDR.
\end{abstract}
%


\begin{highlights}
\item A lightweight multi-scale group convolutional network (MSGC-NetVLAD) for visual place recognition
\item A pedestrian dead-reckoning (PDR) algorithm with magnetic disturbances rejection (MDR)
\item A novel visual/inertial integrated localization scheme with gross error suppression
\item Detailed performance analysis for the MSGC-NetVLAD
\item Detailed performance analysis for the visual/inertial integrated localization scheme in large-scale environments
\end{highlights}



\maketitle

\section{Introduction}\label{Introduction}
Long-term, stable, and continuous pedestrian positioning in large-scale urban environments has been a significant challenge for building smart cities. Although the global navigation satellite system (GNSS) provides reliable and low-cost positioning services, it is easily obstructed by dense buildings or large trees, leading to signal loss and incomplete tracking \cite{gu2015gnss}. For this issue, inertial tracking methods \cite{zhang2021learning,potorti2021off,chen2022deep}, such as pedestrian dead reckoning (PDR), emerge as a promising solution. PDR estimates the user's position by combining step-length estimation with heading information, typically obtained from self-contained sensors such as accelerometers, gyroscopes, and magnetometers. Although PDR is highly useful in GNSS-denied environments and can operate independently of external signals, its primary drawback lies in drift errors that accumulate over time due to sensor noise, incorrect step detection, and inaccurate step length estimation \cite{chen2024deep}.

To mitigate drift errors, researchers have explored the combination of PDR with additional enhancing methods, such as Ultra-Wide-band (UWB), WiFi, Bluetooth, etc. These methods have shown impressive indoor performance. However, they often require additional infrastructure and are unsuitable for large-scale outdoor environments due to their high costs and limited coverage. Fortunately, the emergence of augmented reality (AR) enables devices to obtain visual information in real time for positioning. In existing methods, visual simultaneous localization and mapping (SLAM), visual place recognition (VPR), visual-inertial methods, and more have been proposed. Visual SLAM \cite{mur2015orb,bamdad2024slam} offers a powerful method by using cameras to build a map of the environment and localize the user. Nevertheless, visual SLAM struggles in large-scale, dynamic or poorly lit environments, and its high computational demands pose a challenge for real-time applications on portable devices with limited processing power. 

VPR is typically approached as an image retrieval task \cite{arandjelovic2016netvlad,hausler2021patch,yu2019spatial,xu2021esa}. Given a query image of the target position, the system retrieves the best-matching reference image with a pre-installed position by traversing a database. With the establishment and enrichment of urban geographic databases, such as street view maps, VPR has become a powerful tool to capture opportunistic visual signals for urban localization. These signals refer to the use of occasional, context-dependent visual information—such as landmarks, street views, or fixed visual beacons—to enhance localization systems in environments where traditional GNSS signals are weak or unavailable, such as urban canyons or indoor spaces. These visual signals serve as an alternative positioning method that complements GNSS, allowing for reliable localization even in challenging conditions. The central challenge for VPR is to create a compact image descriptor to represent an image effectively. Among all the methods, the aggregation-based techniques have demonstrated superior performance. Inspired by traditional VLAD \cite{jegou2010aggregating}, NetVLAD \cite{arandjelovic2016netvlad} has emerged as the most prominent global image representation for VPR. In pursuit of even better performance, researchers have enhanced NetVLAD with additional functional modules \cite{peng2021semantic,yu2019spatial,peng2021attentional}. Although these variants achieve state-of-the-art results, they rely on large backbone networks such as VGG16 \cite{simonyan2014very} for feature extraction. The complexity of these deep architectures increases both memory usage and computational costs, placing greater demands on hardware devices. In addition, the VPR algorithm only provides discrete position estimation instead of a continuous trajectory. 

To fully leverage the complementary strengths of inertial and visual technologies, visual-inertial methods such as visual-inertial SLAM \cite{qin2018vins}, and PDR algorithm enhanced by visual re-localization \cite{chen2023reloc} have been proposed. However, these visual-inertial SLAM algorithms are often constrained by portable devices' limited computational power and energy capacity. A dense visual feature map must be pre-installed in the approach described in \cite{chen2023reloc}. Building such a map, however, is challenging in open or poorly lit areas due to the insufficient availability of visual or geometric features. Additionally, a precise map is required and any inaccuracies in mapping can negatively impact positioning error correction, leading to sub-optimal localization performance. Moreover, creating and renewing dense maps require significant computational resources, further taxing device performance.

In this paper, we propose a magnetic disturbances rejection (MDR)-aided PDR algorithm, termed MDR-PDR, complemented by a lightweight VPR neural network, MSGC-Net, to address the aforementioned limitations. Specifically, our contributions are summarized as follows: 
\begin{itemize} 
\item To fully utilize both inertial and magnetic measurements while mitigating magnetic distortion, we introduce the MDR-PDR algorithm, which employs magnetic disturbance rejection for reliable continuous position estimation. 
\item We design a lightweight VPR neural network, MSGC-NetVLAD, that provides absolute position references for the MDR-PDR algorithm. 
\item To ensure reliable long-term positioning, we integrate the MDR-PDR and VPR systems using a Kalman filter which offers trajectory smoothing and gross error suppression (GES). 
\end{itemize}

\section{Methodology}\label{Methodology}
\subsection{Overview}
\begin{figure}[t]
  \centering
   \includegraphics[width=1\linewidth]{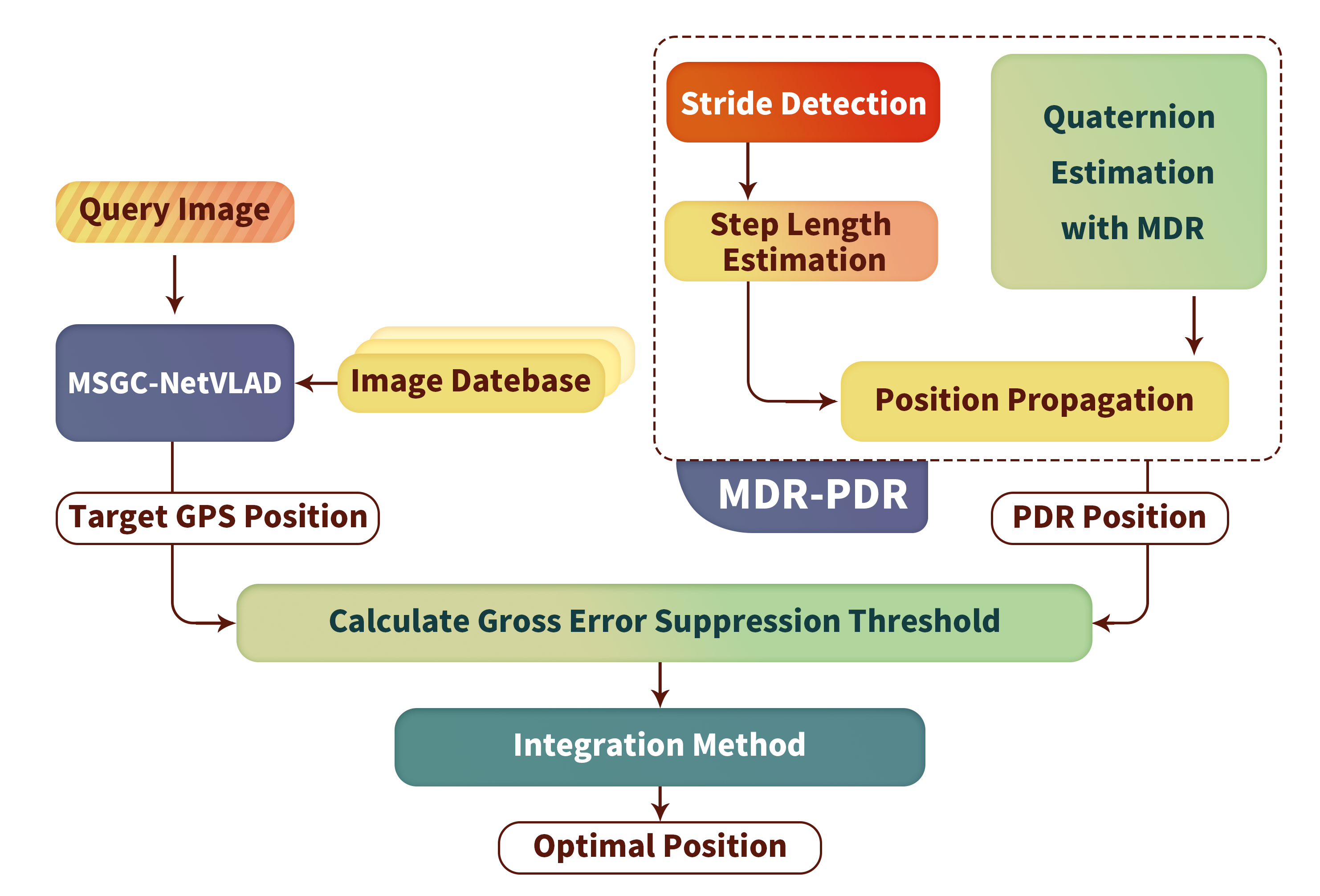}
    \caption{Overview of the proposed PDR-VPR Algorithm}\label{overview}
\end{figure}

Our method utilizes a fine-designed VPR neural network (MSGC-NetVLAD) to recognize opportunistic visual beacons to supplement the PDR method (MDR-PDR), towards low-cost, real-time continuous positioning in GNSS-denied large-scale environments. The architecture of the entire algorithm is shown in Figure \ref{overview}. 

\subsection{MSGC-NetVLAD for VPR}
The process of VPR involves pre-collecting an offline image beacon database. This database consists of a set of images and their corresponding locations, with each image-location pair serving as a beacon. When the user captures an image, the algorithm traverses the database, matches the captured image with the beacons in the database, and assigns the location of the successfully matched beacon as the current location of the user's captured image. Therefore, VPR is essentially an image retrieval problem.

\begin{figure*}[h]
  \centering
   \includegraphics[width=1\linewidth]{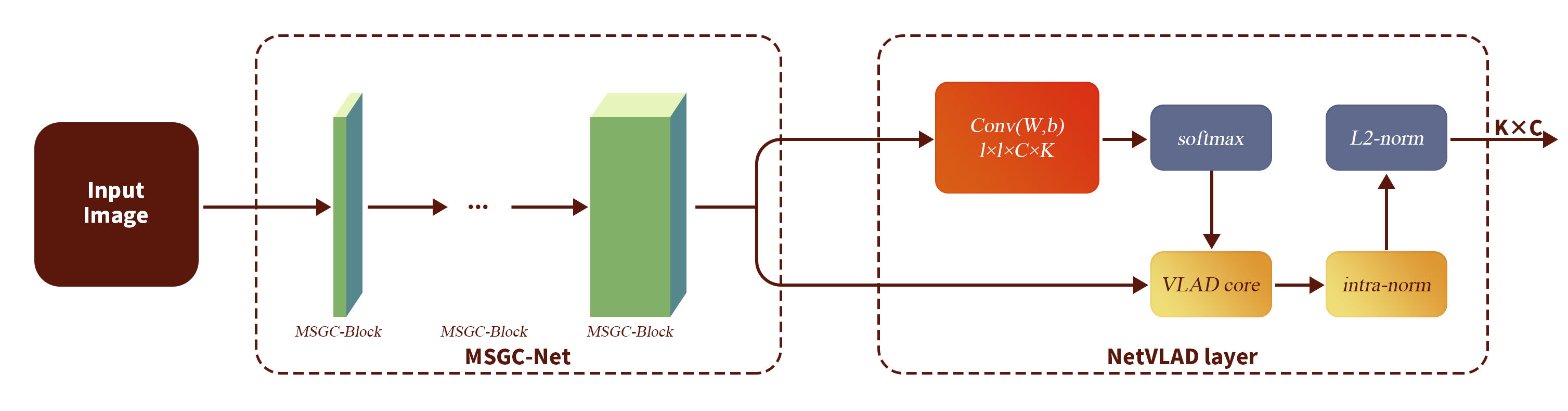}
    \caption{Overview of MSGC-NetVLAD}\label{MSGC-Net}
\end{figure*}

As shown in Figure \ref{MSGC-Net}, the proposed MSGC-Net comprises 4 cascaded MSGC blocks (Section 2.2.3) and a pooling layer. Each MSGC block consists of 2 MSGC modules (Section 2.2.1) and a channel attention module (Section 2.2.2). Each MSGC module contains several dilated convolution kernels with different dilation rates to extract multi-scale information. Finally, a \textit{NetVLAD} serves as a pooling layer to map the last MSGC block's output into the global query image descriptor.

\subsubsection{MSGC Module}
In this section, we will introduce the details of the MSGC module. In our visual beacon recognition task, significant information for the recognizer is usually provided by fixed objects such as buildings. The features of such objects typically manifest at different spatial scales: large scale for building outlines, and small scale for building texture details. 
\begin{equation}
X=\{x_{scale 1},...,x_{scale n},...,x_{scale N}\}
\end{equation}
Feature \(X\) is presented as the combination of different scale of features \(x_{scale n}\). Dilated convolution is a method to adjust the receptive field of the convolution kernel while keeping the number of parameters constant. We use dilated convolution kernels with different dilation rates to extract features at various scales.
\begin{equation}
\hat{X}_D=Dconv(\omega_d,d,X)
\end{equation}
This allows us to adjust the receptive field of the convolution kernel by changing the dilation rate \(d\) without increasing the number of parameters. To achieve a lightweight design, we introduce Group Convolution to reduce the number of parameters further \cite{huang2018condensenet}.
\begin{equation}
\hat{X}_G=Cat\{Gconv(\omega_g,X^{(1)}),...,  \\
Gconv(\omega,X^{(g)})\}
\end{equation}
Group convolution divides the feature map into \(g\) groups based on channels, with each group sharing parameters internally to achieve efficient parallel convolution operations. The parameters of the group convolution will be reduced to \(1/g\) of those in the original convolution. Our MSGC module combines group convolution and dilated convolution, we apply dilated group convolution with multiple dilation rates \(d\) to the feature maps in parallel, as shown in the following formula:
\begin{equation}
\begin{split}
    \hat{X}_{GD}=\sum_{d}^{D}Cat\{GDconv(\omega_{d,g},d, X^{(1)}),...,
    \\GDconv(\omega_{d,g},d, X^{(g)})\}
\end{split}
\end{equation}
where \(d\in D\), \(D\) is the set of different dilation rates. We combine the features from different dilation rates using summation, which helps to avoid excessive channel expansion and further reduce computational complexity. It has been proven in [ref group] that grouped convolution reduces computational complexity at the cost of losing some information. To retain sufficient information, we adopt a method similar to residual connections to supplement the feature maps, That is, point-wise convolution is used to preserve the information from the previous layer's feature map. This method employs a 1x1 convolution kernel to enhance the nonlinear representation of the feature map without altering its spatial information:
\begin{equation}
\hat{X}_{P}=conv_{1x1}(\omega_p,X)
\end{equation}
We concatenate the original spatial features and multi-scale spatial features along the channel dimension:
\begin{equation}
\hat{X}=Cat\{\hat{X}_{P},\hat{X}_{GD}\}
\end{equation}
In summary, we present the output \(\hat{X}\) of the input feature map \(X\) after being processed by an MSGC module.

\subsubsection{Scale Attention Module}
In the previous subsection, we introduced the basic MSGC module, where we concatenated the original scale features and multi-scale grouped convolution features along the channels. In this subsection, we use channel attention to weight the feature maps, indicating the varying contributions of different features to visual beacon recognition. We introduced the ECA module [\cite{wang2020eca}] to implement channel attention, as equation(\ref{ECA}):
\begin{equation}
\label{ECA}
    \omega_c=sigmoid(conv_{1d}(GAP(\hat{X})))
\end{equation}
Global Average Pooling(GAP) is used to aggregate the feature information of each channel into a single element. The entire feature map is transformed to a $1xC$ vector, where $C$ is the number of channels. Afterward, a 1D convolution is used to learn the interactions between channels, and the $sigmoid$ function is applied to map the output to a range of 0 to 1, serving as the weights for the channels.

\subsubsection{MSGC Blocks}
\begin{figure}[t]
  \centering
   \includegraphics[width=1\linewidth]{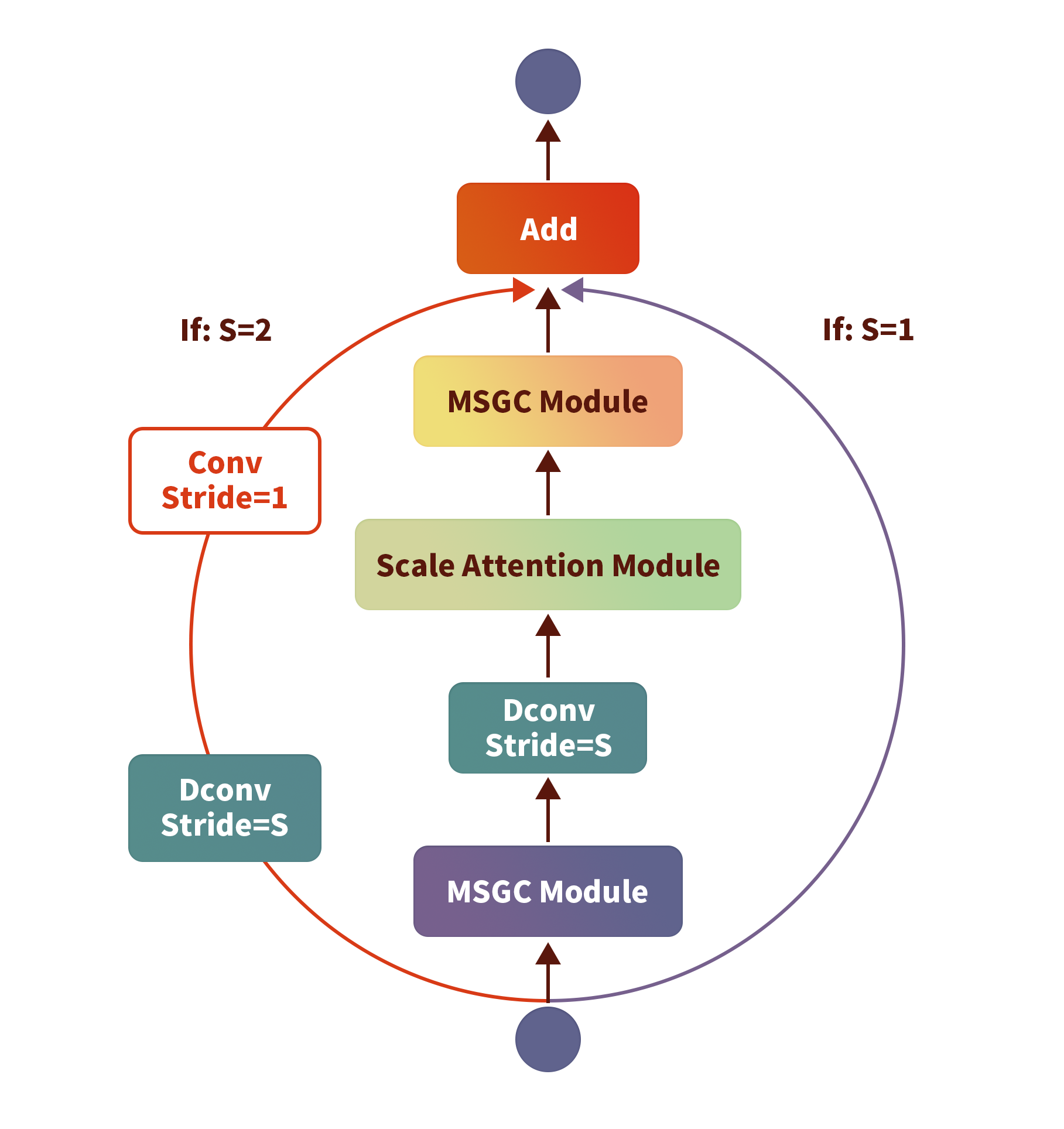}
    \caption{Optional Structure for MSGC-Block}\label{MSGC-Block}
\end{figure}

In this subsection, we stack MSGC modules into an MSGC block, using a bottleneck approach to achieve spatial compression and dimensional expansion within the block. This method has been shown in [ref ResNet] to enhance feature representation while reducing computational complexity. As shown in Figure \ref{MSGC-Block}, we implement dimensional expansion within the MSGC module and add a grouped convolution layer between two MSGC modules to achieve spatial compression. Additionally, we include a residual connection within the MSGC block to prevent the vanishing gradient problem that can occur in deep neural networks.

The MSGC-Block has two parameter options: a parameter $S$ of 1 or 2. When the $S$ is set to 1, the spatial dimensions of the features remain unchanged, while with a $S$ of 2, the spatial dimensions are compressed by half. In our network architecture, we alternate between these two configurations of the MSGC-Block. Detailed implementation specifics are introduced in the experimental section.

\subsubsection{NetVLAD Layer and Reference Position Output}
After the MSGC encoding architecture, we add a trainable NetVLAD pooling layer \cite{arandjelovic2016netvlad} to map the extracted multi-scale local features into global descriptors. NetVLAD supports the use of deep features to replace traditional VLAD, as shown in Figure \ref{MSGC-Net}.

The VLAD layer has several cluster centers that are learned during training. Each deep feature is assigned to one or more of these clusters. For each feature, NetVLAD calculates the residuals, the differences between the feature and the cluster centers it is assigned to. These residuals capture how the features deviate from their assigned cluster centers. NetVLAD aggregates these residuals for all features of the same cluster, summing up the residuals to form a descriptor for each cluster. This results in a descriptor that summarizes the feature variations relative to each cluster. The resulting aggregated descriptor is then L2-normalization used to ensure consistent magnitude, resulting in a compact global descriptor that represents the entire image.

The similarities between the global descriptors of a query image and those in the image database are computed to generate a ranked similarity list of matches, each associated with absolute position labels in descending order. We select the top \textbf{25} entries from this list and apply a quality control step based on the consistency of the matched image locations. Rather than fully relying on the image with the highest similarity, we choose the image whose location appears most frequently among the top \textbf{25} matches. This location is then used as the reference position.

\subsection{Dead Reckoning Method}
The pipeline of our proposed MDR-PDR algorithm consists of the following stages:
\subsubsection{Stride Detection}
PDR algorithm rely heavily on precise stride detection to estimate user displacement during movements. Stride detection is commonly achieved by analyzing the accelerometer readings. One major challenge in such algorithm is the suppression of noise and the accurate identification of steps amidst random signal fluctuations. To address this, we propose an enhanced step detection approach that leverages the magnitude of acceleration signals and enforces a minimum time interval between consecutive detections. This approach enhances accuracy by filtering out spurious peaks that do not correspond to genuine steps.

\textbf{Data pre-processing:} The algorithm first acquires raw acceleration data from the three-axis accelerometer. Given the three-axis acceleration signal \( \boldsymbol{a}_k = [a_k(x) , a_k(y) , a_k(z)]\) at time \(k\), the magnitude of the acceleration is calculated by 
\begin{equation}
\lVert \boldsymbol{a}_k \rVert = \sqrt{a_k^2(x) + a_k^2(y) + a_k^2(z)}
\end{equation}
We apply a sliding window mean filter with a fixed window size \(N\) over the acceleration magnitude to reduce noise and smooth out the signal, then
\begin{equation}
\overline{\lVert\boldsymbol{a}_k \rVert}=\frac{1}{N}\sum_{i = 0}^{N - 1}\lVert\boldsymbol{a}_{k - i}\rVert
\end{equation}

\textbf{Peak Detection with Minimum Time Interval:} First, to identify step events, peaks in the filtered acceleration magnitude must exceed the threshold, i.e., \(\overline{\lVert\boldsymbol{a}_p \rVert}>T_{peak}\),
where $T_{peak}$ is a constant to identify standstill and movement. In addition, the interval between two consecutive peaks must exceed a minimum threshold \(T_{time}\) typically set to match the expected time between steps, i.e., \(t_p - t_{p - 1}>T_{\text{time}}\). This time constraint helps mitigate false positives from transient movements or noise in the accelerometer readings. Therefore, mathematically, a step event meets
\begin{equation}
\overline{\lVert\boldsymbol{a}_p \rVert}>T_{peak} \text{ and } t_p - t_{p - 1}>T_{\text{time}}
\end{equation}

\subsubsection{Step Length Estimation}
After the stride detection, step length is updated at a valid step peak by the typical Weinberg step length formula \cite{weinberg2002using}.
\begin{equation}
\label{Weinberg}
L = K \sqrt[4]{acc_{max}-acc_{min}}
\end{equation}
where \(L\) is the estimated step length; \(K\) is a constant (calibrated based on the walking patterns of different users); \(acc_{max}\) and \(acc_{min}\) refer to maximum and minimum vertical accelerations between two consecutive valid peaks.

\subsubsection{Quaternion Estimation with Magnetic Disturbances Rejection}
The proposed PDR algorithm estimates the orientation based on inertial and magnetic sensors using a gradient descent algorithm \cite{madgwick2011estimation}. The quaternion iteration formula is 
\begin{equation}
\label{ori}
\dot{\boldsymbol{q}}_{est,k}=\frac{1}{2} \boldsymbol{q}_{\omega, k} \otimes \boldsymbol{\omega}_k-\beta \nabla \boldsymbol{f}_{a,m}
\end{equation}
where ${\boldsymbol{q}}_{est,k}$ is the estimated quaternion at time $k$; $\boldsymbol{q}_{\omega, k}$ is the orientation measured by the gyroscopes at time $k$ and $\boldsymbol{\omega}_k$ is the angular rate; $\beta$ is a factor that controls the sensitivity of the orientation error correction; $\nabla \boldsymbol{f}_{a,m}$ is the gradient descent step used to minimize the error between the estimated direction of gravity and magnetic field and the actual measurements from the accelerometer and magnetometer.

However, magnetic interference, including electrical appliances, furniture, and metal structures, easily affects the magnetometer readings. Therefore, a detector is employed to identify the interference. Namely, a pure magnetic field is determined when the magnetic field magnitude $\boldsymbol{m}$ and magnetic inclination $I$ are in certain thresholds. Since the mean values provide useful estimates of central tendencies in distribution and can effectively reduce noise, the detector is defined as follows
\begin{equation}
\alpha= \begin{cases}1 & \text {if } \overline{\lVert\boldsymbol{m}_k\rVert}<T_{m} \text { and } I<T_I \\ 0 & \text { otherwise }\end{cases}
\end{equation}
where $0$ and $1$ refer to magnetic disturbance and pure magnetic field respectively;  $\overline{\lVert\boldsymbol{m}_k\rVert}=\frac{1}{N_m} \sum_{i=0}^{N_m-1} \overline{\lVert\boldsymbol{m}_{k-i}\rVert}$ and $ I=cos^{-1}\left(\frac{\bar{\boldsymbol{m}}\left(k-N_m+1: k\right) \cdot \boldsymbol{g}_e}{\lVert\bar{\boldsymbol{m}}\left(k-N_m+1: k\right)\rVert}\right)$ where $\bar{\boldsymbol{m}}\left(k-N_m+1: k\right)$ refers to the mean magnetic field vector in a window with a size of $N_m$ and $\boldsymbol{g}_e$ is the normalized gravity vector. When we do a dot product between the magnetic field vector and a unit gravity vector, we will have the vertical magnetic field strength. $I$ is the angle between the vertical component and the total magnetic field vector.

Therefore, the equation (\ref{ori}) is rewritten as
\begin{equation}
\label{IMadgwick}
\dot{\boldsymbol{q}}_{est,k}=\frac{1}{2} \boldsymbol{q}_{\omega, k} \otimes \boldsymbol{\omega}_k-\alpha \beta \nabla \boldsymbol{f}_{a,m}
\end{equation}
Namely, the orientation is updated only using gyroscope readings in the environment with magnetic disturbances; otherwise, gravity and magnetic field vectors are used to correct the orientation errors. 

\subsubsection{Position Propagation}
The estimated quaternion can be converted to Euler angles (i.e. roll, pitch, and yaw) \cite{diebel2006representing}. Given the yaw ($\psi_k$) at time $k$, the position is updated by
\begin{equation}
\mathbf{p}_{i}=\mathbf{p}_{i-1}+L_i \left[\begin{array}{l}
\cos \left(\psi_i\right) \\
\sin \left(\psi_i\right)
\end{array}\right]
\end{equation}
where $\mathbf{p}_i=\left[x_i, y_i\right]^\text{T}$ is the 2D position at $i$-th step; $L_i$ is the step length computed by equation (\ref{Weinberg}) at $i$-th step; $\psi_i$ is the $i$-th heading converted by quaternion iteration using equation (\ref{IMadgwick}) \cite{diebel2006representing}.

\subsection{Integration Method with Gross Error Suppression}


The gross error issue in our method will be discussed first. In real-world applications, leveraging distinct features such as buildings, landmarks, signs, and other specific identifiers is common for place recognition. However, for the VPR task, we cannot assume that users will always capture unique surrounding features, as images often include common elements like trees, roads, and the sky. Generally, if the query image has a sufficient number of distinct features, the MSGC network can provide reasonably reliable matching results. Nonetheless, when common features dominate the image feature set, the matching results are viewed as unreliable, leading to uncontrollable gross errors. To address this, we suggest utilizing the PDR algorithm as prior information to assess the confidence level of VPR.

\par

The errors in PDR originate from two main sources. The first is attitude error, which includes errors from the accelerometer (its noise satisfies \(n_a\sim N(0,\sigma_a^2)\)), gyroscope (its noise satisfies \(n_{gy}\sim N(0,\sigma_{gy}^2)\)), and magnetometer (its noise satisfies \(n_m\sim N(0,\sigma_m^2)\)). The second source of error comes from the empirical step length model. In our attitude sensing algorithm, the attitude estimation error \(\delta q_{gy}\) obtained from the gyroscope is:
\begin{equation}
    \delta q_{gy}=\frac{1}{2}q_k\otimes(n_{gy})\Delta t
\end{equation}
where \(q_k\) represents the attitude quaternion at the current time, and in the process of reference direction optimization based on the accelerometer and magnetometer, the corresponding estimation error is:
\begin{equation}
    \delta q_{a,m}=-\beta \nabla{q_k} (q_{gy}, a_k, m_k)
\end{equation}
where \(\nabla{q_k}\) is the gradient function of the current attitude quaternion. Then the variance corresponding to this estimated error is:
\begin{equation}
    \sigma^2_{\delta q_{k+1}} = \frac{1}{4}  q_k \otimes (\sigma_{gy}^2)\Delta t^2 + \alpha \beta^2(\sigma_{a}^2 +\sigma_{m}^2)
\end{equation}

To calculate the position estimation error, it is important first to determine the stride length estimation error. Given that we utilize the empirical model to estimate stride length, analyzing how sensor noise propagates to affect step length estimation can be complex. As a result, we rely on the relative accuracy reported in the literature to define the step length estimation error. From Equation (\ref{IMadgwick}) we can derive, in a single step, the propagation of position error can be expressed as:
\begin{equation}
    \Vert \delta \mathbf{p}_{i+1}\Vert= \sqrt{\delta L_i^2 + (L_i\delta\psi_i)^2}
\end{equation}
where \(\delta L_i\) represents the stride length error of the time of $i$-th step, and \(\delta\psi_i\) denotes the heading angle error of the time of $i$-th step, determined by the attitude error \(\delta q_{i}\) of that step. The variance of the magnitude of the position for the time of $i$-th step can be expressed as:
\begin{equation}
    \sigma_{\Vert \delta \mathbf{p}_{i+1}\Vert}^2= \sigma^{2}_{L}+\sigma_{\delta \psi_i}^2L_i^2 + \sigma_{\Vert \delta \mathbf{p}_{i}\Vert}^2
\end{equation}
where \(\sigma_L\) is the standard deviation of the stride length estimation, which is reported in the literature as 15$\%$ of the true step length \cite{weinberg2002using}. Adjacent visual landmarks may be confused for the VPR algorithm due to similar geographical features. To address this, a margin $\gamma$ is required to distinguish visual beacons with close positions. Its value is generally set to half of the distance between adjacent points. Therefore, the threshold $T$ can be defined as the sum of the accumulated standard deviation of the total position drift error and a constant $\gamma$:
\begin{equation}
    T = \sqrt{\sum_{i=1}^N \sigma^2_{\Vert \delta \mathbf{p}_{i}\Vert}}+\gamma
\end{equation}


We have designed a Kalman filter to integrate the VPR and PDR. In VPR tasks, GPS positions are also provided. Then the query image coordinates serve as observations for the Kalman filter. However, the VPR neural network has performance limitations that prevent us from completely relying on its output, particularly in regions with similar geographical features. To address this, we implement the threshold $T$ for conditional observation. If the Euclidean distance between the PDR and VPR outputs exceeds $T$, this suggests that the error in the VPR system is greater than the natural drift error in PDR. In such cases, it is more prudent to rely more on the PDR data. Conversely, if the distance is within the threshold, we place more trust in the VPR system.


The state vector of the proposed Kalman filter is represented as \(\mathbf{x} = \mathbf{p}\), where \(\mathbf{p}\) denotes the position. During the prediction phase, the position vector \(\mathbf{p}\) is determined by propagating the position from the PDR system. In the event of a step, we utilize the positions of the top 25 images from the VPR to calculate the Euclidean distances relative to the PDR. The update step is initiated when one of the distances falls below the threshold \(T\). The VPR position denoted as \(\mathbf{z}\) is then used to compute the measurement residual \(\mathbf{y}\). Since the position correction may cause a sudden change, a low-pass filter is used to smooth the results. Ultimately, the filter outputs the fused position. This entire process is detailed in Algorithm 1.

\begin{algorithm} 
\caption{Integration Kalman Filter Design}
\KwIn{

Initial state vector: $\mathbf{x}_0=\mathbf{p}_0$ \\
Initial state covariance: $\mathbf{P}_0$ \\
State transition matrix: $\mathbf{A}=\mathbf{I}_{2}$
 \\
Observation matrix: $\mathbf{H}=\mathbf{I}_{2}$
 \\
Process noise covariance: $\mathbf{Q}$ \\
Measurement noise covariance: $\mathbf{R}$
}
\KwOut{Estimated state $\mathbf{x}_k$ and covariance $\mathbf{P}_k$ at each step.}

\BlankLine
Set initial state estimate: $\mathbf{x}_0$ \\
Set initial covariance: $\mathbf{P}_0$

\For{each step $i = 1, 2, \dots$}{
    \textbf{Prediction Phase:}
    \begin{itemize}
        \item Predict the PDR state: $\mathbf{{x}}_i = \mathbf{A} \mathbf{x}_{i-1} +L_i \left[\begin{array}{l}
\cos \left(\psi_i\right) \\
\sin \left(\psi_i\right)
\end{array}\right]$

        \item Predict the covariance: $\mathbf{{P}}_i = \mathbf{A} \mathbf{P}_{i-1} \mathbf{A}^\top + \mathbf{Q}$

        \item Calculate the Euclidean distances between the top-25 positions from the VPR algorithm and the current PDR position $\mathbf{{x}}_i$

    \end{itemize}

    \textbf{Update Phase (If one of the distances falls below the gross error suppression threshold \(T\)):}
    \begin{itemize}
       \item Record the VPR position $\mathbf{z}_i$ that satisfies the gross error suppression threshold in the top-25 list
    
        \item Compute the measurement residual (innovation): 
        $
        \mathbf{y}_i = \mathbf{z}_i - \mathbf{H} \mathbf{{x}}_i
        $
        \item Compute the innovation covariance: 
        $
        \mathbf{S}_i = \mathbf{H} \mathbf{{P}}_i \mathbf{H}^\top + \mathbf{R}
        $
        \item Compute the Kalman gain: 
        $
        \mathbf{K}_i = \mathbf{{P}}_i \mathbf{H}^\top \mathbf{S}_i^{-1}
        $
        \item Update the fused state estimate: 
        $
        \mathbf{\hat{x}}_i = \mathbf{{x}}_i + \mathbf{K}_i \mathbf{y}_i
        $
        \item Update the covariance estimate: 
        $
        \mathbf{\hat{P}}_i = (\mathbf{I} - \mathbf{K}_i \mathbf{H}) \mathbf{{P}}_i
        $

        \item Smooth the trajectory $\mathbf{\hat{x}}_i=a\mathbf{\hat{x}}_i+(1-a)\mathbf{x}_{i-1}$

    \end{itemize}
    
}

\Return{$\mathbf{x}_i$, $\mathbf{P}_i$}
\end{algorithm}

\section{Experimental Results and Analysis}
We adopt two commonly used public datasets and a private dataset to evaluate our algorithm, focusing on the following aspects: performance evaluation of VPR, including performance comparisons with other lightweight models; also evaluation of the complexity and parameter count of the neural networks; evaluation of the MDR-PDR algorithm; evaluation of the entire localization algorithm, including the accuracy and continuity of positioning trajectories.

\subsection{Dataset and Evaluation Metric}
We evaluated our proposed method using the Pitts30k \cite{pitts30k}, Tokyo 24/7 \cite{torii201524}, and a self-collected Real-World Walk dataset. The Pitts30k and Tokyo 24/7 datasets are widely utilized for evaluating VPR. The Pitts30k dataset consists of three parts: a training set, a validation set, and a test set, with each containing 10,000 image-location pairs. In total, it includes 30,000 images sourced from Google Street View and 22,000 test queries captured at various times. 

The Tokyo 24/7 dataset is a public test dataset that comprises 76,000 database images and 315 query images taken with mobile phone cameras. This dataset presents a significant challenge, as the queries were collected during different times of the day—daytime, sunset, and nighttime—while the database images were exclusively captured during the daytime using Google Street View.

Additionally, we employed the Places365 dataset \cite{places365}, which contains a collection of building images with 365 distinct labels and approximately 1.8 million image-building label pairs in its standard version. We believe that buildings contain some of the most important location-related features for VPR tasks, which is why we pretrain our model using the Places365 dataset. After completing the pretraining for building image classification on this dataset, we further train the VPR models specifically on the Pitts30k training dataset. Finally, we use the Pitts30k-test dataset and the Tokyo 24/7 dataset to evaluate the model's performance.

For the final evaluation of the proposed visual-inertial localization algorithm, we incorporated the Real-World Walk dataset, which includes two trajectories with time-aligned IMU data, discrete images, and ground truth positions. Additionally, 33 visual landmarks were employed as the VPR database.

Consistent with standard evaluation protocols for these datasets, we measured VPR model performance using Recall@N \cite{arandjelovic2016netvlad}. Additionally, the efficiency of the proposed VPR model was assessed through model parameter quantity and floating-point operations per second (FLOPS). GPS benchmarks were employed to evaluate the efficacy of our visual-inertial localization algorithm.

\subsection{VPR Neural Network Implementation Details}
In this work, all experiments for neural networks are conducted in PyTorch. VGG-16 \cite{simonyan2014very}, AlexNet \cite{krizhevsky2012imagenet}, MobileNetV3 \cite{howard2017mobilenets} and GhostNet \cite{han2020ghostnet} are selected as the optional backbone encoding networks for comparison experiments, where MobileNetV3 and GhostNet are the typical lightweight neural networks. AlexNet and VGG-16 are cropped at the last convolution layer with the encoder dimension of
256 and 512 before the ReLU layer, respectively. MobileNetV3 and GhostNet are cropped at the last stage with the encoder dimension of 960 before the average pooling layer. Following image representation learning for retrieval-based VPR \cite{arandjelovic2016netvlad}, we use the SGD optimizer to minimize the triplet ranking loss for tuple metric learning. Given a query image $q$, a triplet tuple is defined as $\left(q,\left\{p_{i}^{q}\right\},\left\{n_{j}^{q}\right\}\right)$, where $\left\{p_{i}^{q}\right\}$ is a set of potential positives, the smallest descriptor
distance to the query, and $\left\{n_{j}^{q}\right\}$ is a set of definite negatives. We need to choose the best positive ($p_{i *}^q={\operatorname{argmin}} d_\theta\left(q, p_i^q\right)$). Namely, for a given query image $q$, we wish that the Euclidean distance between the query $q$ and the best potential positive to be smaller than its distance to the definite negative, which is described as
\begin{equation}
\label{eq6}
\mathop{d_{\theta}\left(q, p_{i *}^{q}\right)<d_{\theta}\left(q, n_{j}^{q}\right), \forall j}
\end{equation}
Thus, the triplet ranking loss $L_{\theta}$ is described as:
\begin{equation}
\label{eq7}
\mathop{L_{\theta}=\sum_{j} l\left(\min _{i}\left(  d_{\theta}^{2}\left(q, p_{i}^{q}\right)\right) +m-d_{\theta}^{2}\left(q, n_{j}^{q}\right)\right)}
\end{equation}
where $m$ is a constant margin as 0.1 to ensure that the query is close to the potential positive and away from the definite negative; $l$ is the hinge loss $l\left(x \right)=max\left(0,x \right)  $.
All models are trained on the Pitts30k-train dataset following the same pipeline in \cite{arandjelovic2016netvlad}. 

\subsection{VPR Performance Evaluation}
The MSGC-Net is only trained on the Pitts30k-train dataset, with its performance evaluated on the Pitts30k-test dataset. In order to further assess the model’s scalability, we also assessed its performance on the Tokyo24/7 dataset, alongside the classic VGG16-based and several commonly used lightweight models.

\begin{table}[t]
\caption{FLOPS and number of trainable parameters of different models. We report all results, including the \textcolor{red}{best} and \textcolor{blue}{second-best} outcomes.}\label{flops and params}
\begin{tabular*}{8cm}{ccc}
\toprule
         Methods &FLOPS (M)& \makecell{Trainable \\ Params.(M)} \\
\midrule         
        VGG16-NetVLAD& 15353.05& 14.75\\
        AlexNet-NetVLAD& 658.34 & \textcolor{blue}{2.49} \\
        MobileNetV3-NetVLAD& 235.29 & 3.03  \\
        GhostNet-NetVLAD& \textcolor{red}{156.31} & 2.73   \\
        MSGC-NetVLAD (Ours) & \textcolor{blue}{181.24} & \textcolor{red}{1.11}  \\
\bottomrule
\end{tabular*}
\end{table}

We initially recorded the FLOPS and the number of trainable parameters for different models. As presented in Table \ref{flops and params}, our method demonstrates a remarkable reduction in parameters—tens of times fewer—compared to the classic benchmark VGG16 within the NetVLAD framework. Additionally, it shows an impressive reduction of nearly a hundred times in terms of FLOPS. In comparison to other lightweight models, our method offers approximately twice the advantage in terms of parameter reduction. In relation to MobileNetV3, which emphasizes computational real-time performance, our method achieves a comparable number of FLOPS while reducing the parameter count by nearly two times.

\begin{table*}[t]
\caption{Performance on Pitts30k-test Dataset. We report all results, including the \textcolor{red}{best} and \textcolor{blue}{second-best} outcomes.}\label{tab: Pitts30k}
\begin{tabular*}{12.1cm}{ccccccc}
\toprule
        \multirow{2}{*}{\centering Methods} & \multicolumn{5}{c}{Pitts30k-test}\\
         \cmidrule(lr){2-6}
         & Recall@1 &Recall@5 &Recall@10 &Recall@20 &Recall@25  \\
\midrule         
        VGG16-NetVLAD& \textcolor{blue}{81.47$\%$} & \textcolor{blue}{90.98$\%$} & \textcolor{red}{93.71$\%$} & \textcolor{red}{95.38$\%$} & \textcolor{red}{96.04$\%$} \\

        AlexNet-NetVLAD& 69.88$\%$  & 85.04$\%$ & 89.14$\%$ & 92.53$\%$ & 93.79$\%$ \\

        MobileNetV3-NetVLAD& 79.78$\%$ & 89.73$\%$ & 92.46$\%$ & 94.57$\%$ & 95.29$\%$ \\

        GhostNet-NetVLAD& 75.43$\%$  & 89.26$\%$ & 92.55$\%$ & 94.78$\%$ & 95.33$\%$ \\

        MSGC-NetVLAD (Ours) & \textcolor{red}{82.78$\%$} & \textcolor{red}{91.40$\%$} & \textcolor{blue}{93.52$\%$} & \textcolor{blue}{95.19$\%$} & \textcolor{blue}{95.82$\%$} \\

\bottomrule
\end{tabular*}
\end{table*}

\begin{table*}[t]
\centering
\caption{Performance on Tokyo 24/7 Dataset. We report all results, including the \textcolor{red}{best} and \textcolor{blue}{second-best} outcomes.}\label{tab: Tokyo247}
\begin{tabular*}{12.1cm}{ccccccc}
\toprule
         \multirow{2}{*}{\centering Methods} & \multicolumn{5}{c}{Tokyo 24/7}\\
         \cmidrule(lr){2-6}
         & Recall@1 &Recall@5 &Recall@10 &Recall@20 &Recall@25  \\
\midrule         
        VGG16-NetVLAD& \textcolor{red}{61.27$\%$} & \textcolor{red}{77.14$\%$}& \textcolor{red}{83.49$\%$} & \textcolor{red}{86.67$\%$} & \textcolor{red}{87.94$\%$} \\

        AlexNet-NetVLAD & 38.10$\%$  & 54.92$\%$ & 60.32$\%$ & 67.30$\%$ & 69.52$\%$ \\

        MobileNetV3-NetVLAD& 44.13$\%$ & 53.97$\%$ & 59.37$\%$ & 64.13$\%$ & 65.71$\%$ \\

        GhostNet-NetVLAD& 37.14$\%$  & 50.16$\%$ & 55.56$\%$ & 60.32$\%$ & 61.27$\%$ \\

        MSGC-NetVLAD (Ours) & \textcolor{blue}{48.89$\%$} & \textcolor{blue}{61.27$\%$} & \textcolor{blue}{66.03$\%$} & \textcolor{blue}{70.16$\%$} & \textcolor{blue}{72.38$\%$} \\
        
\bottomrule
\end{tabular*}
\end{table*}

In our performance evaluations conducted on the Pitts30k-test dataset, detailed in Table \ref{tab: Pitts30k}, we made adjustments to the VGG16 model by freezing some of its layers to minimize its large parameter count. However, even with this modification, VGG16's parameter count remained notably higher than that of the other models. While VGG16 retains a slight performance advantage of less than 0.5\%, this is accompanied by a computational complexity that is several times greater. On the other hand, our proposed MSGC-NetVLAD has demonstrated superior performance in comparison to other methods, particularly excelling in Recall@1, where it showcases a significant advantage. In terms of computational complexity, MSGC-NetVLAD also achieves a noteworthy reduction in parameter numbers.

Furthermore, we assessed and compared the performance of all models on the Tokyo24/7 dataset, as illustrated in Table \ref{tab: Tokyo247}. In contrast to the findings on the Pitts30k-test dataset, VGG16 displayed a more pronounced performance gap on the Tokyo24/7 dataset. This observation may be attributed to its higher parameter count and complexity. Nevertheless, MSGC-NetVLAD continues to demonstrate a clear advantage relative to other lightweight models.

\subsection{PDR-VPR Localization Performance Evaluation}
\begin{figure*}[t]
  \centering
   \includegraphics[width=1\linewidth]{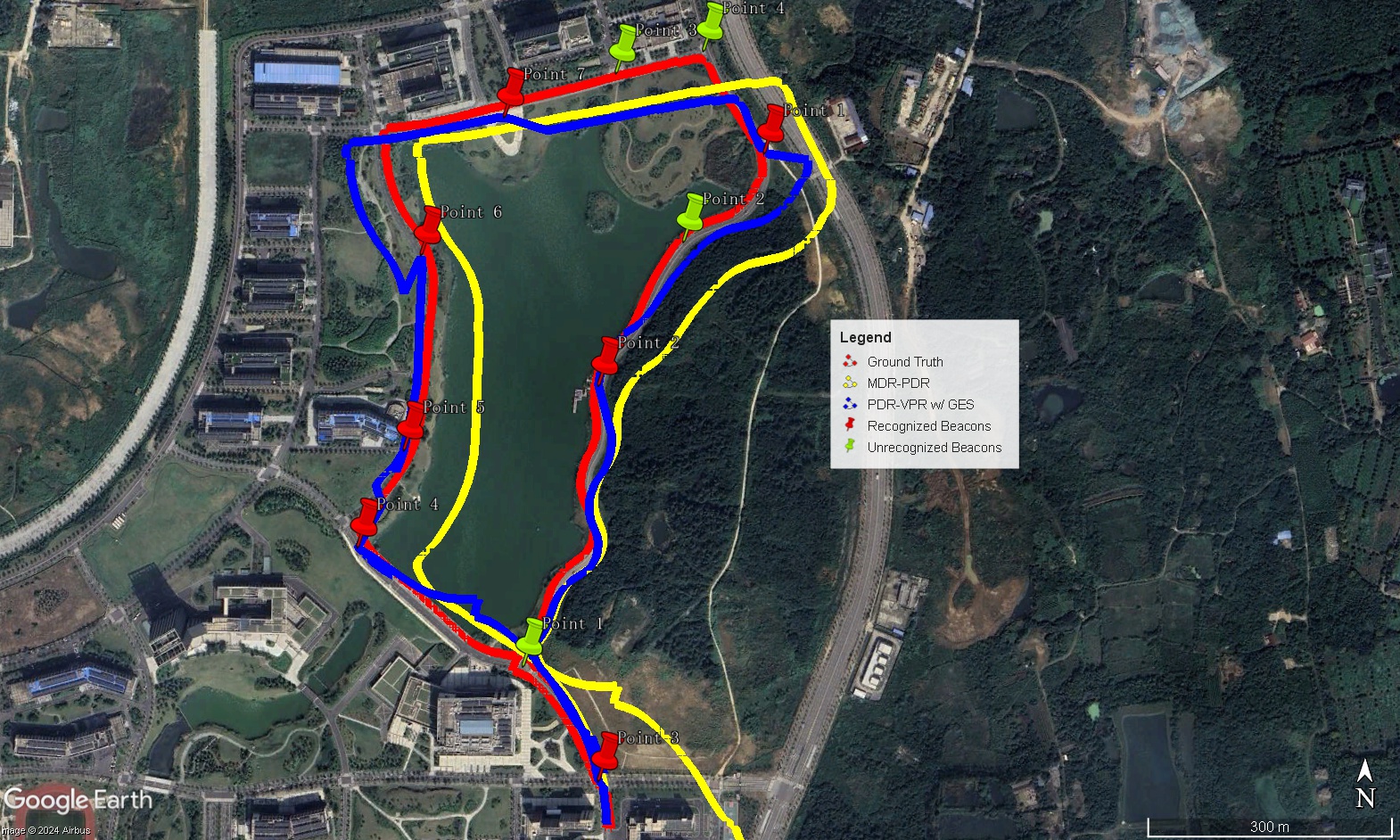}
    \caption{Trajectory comparison of Trajectory 1 in a sparse feature environment. The red points are correctly recognized, while the green ones are unrecognized. The indices of the green points from bottom to top are 1, 3, and 4. PDR-VPR w/ GES refers to the proposed visual-inertial localization algorithm; MDR-PDR represents the proposed dead-reckoning method.}\label{0827_traj}
\end{figure*}

\begin{figure*}[t]
  \centering
   \includegraphics[width=1\linewidth]{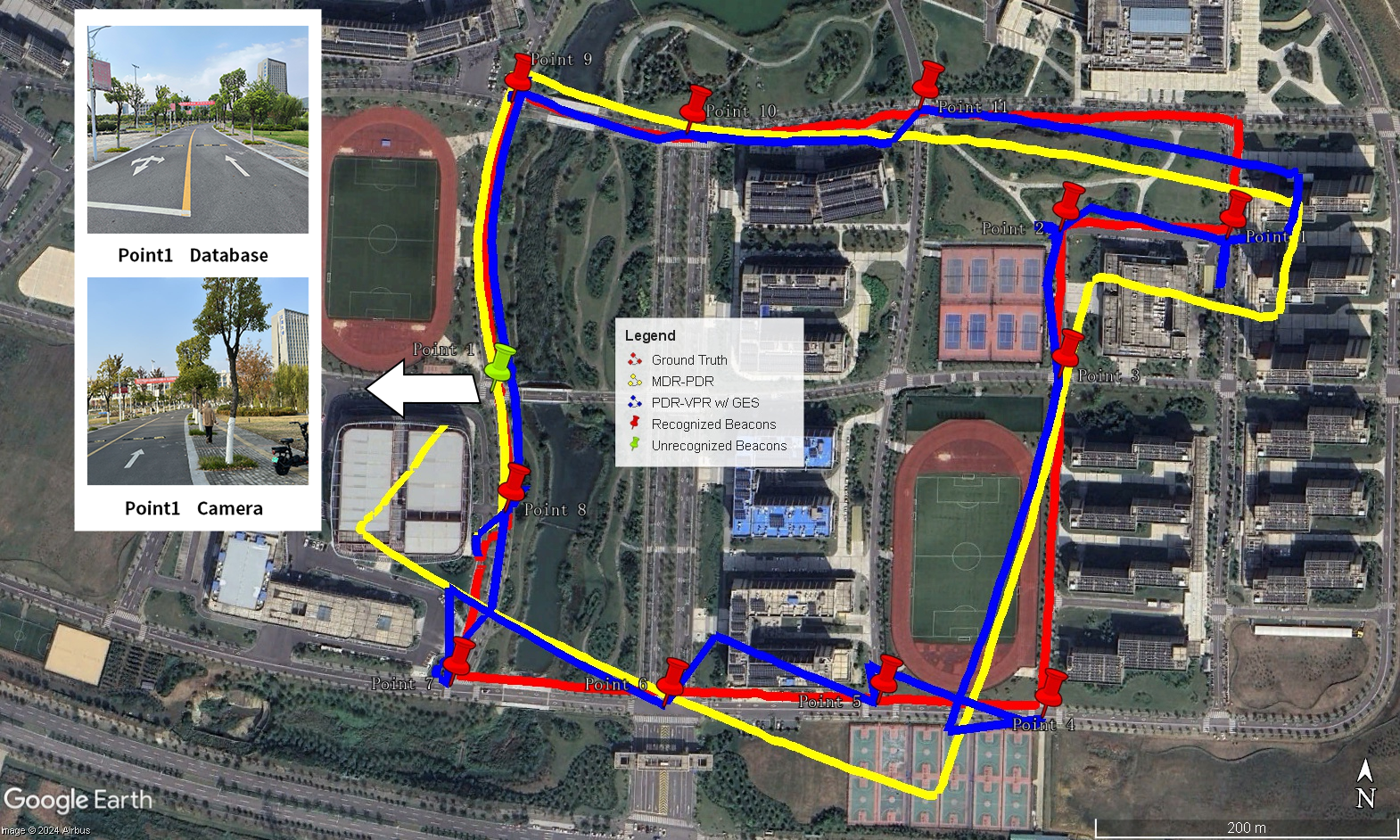}
    \caption{Trajectory comparison of Trajectory 2 in dense feature environment. The red points are correctly recognized, while the green ones are unrecognized. The index of the green point is 1. PDR-VPR w/ GES refers to the proposed visual-inertial localization algorithm; MDR-PDR represents the proposed dead-reckoning method.}\label{1112_traj}
\end{figure*}

We evaluate the PDR-VPR integration algorithm on two long-term (approximately 30 minutes) Real-World Walk trajectories, with participants maintaining a natural walking posture and looking straight ahead. The two trajectories were performed in different visual environments:
\begin{itemize} 
\item Trajectory 1, Sparse Feature Environment: This took place in a typical low-feature and repetitive-feature environment, where the participant walked a closed loop along a lakeside, as shown in Figure \ref{0827_traj}. During the walk, the environment included sparse buildings and few available visual features.
\item Trajectory 2, Dense Feature Environment: This was conducted in a typical high-feature environment (see Figure \ref{1112_traj}), where the participant walked a closed loop among densely packed buildings. The environment offered numerous available visual features for positioning.
\end{itemize}

\begin{figure*}[t]
  \centering
   \includegraphics[width=1\linewidth]{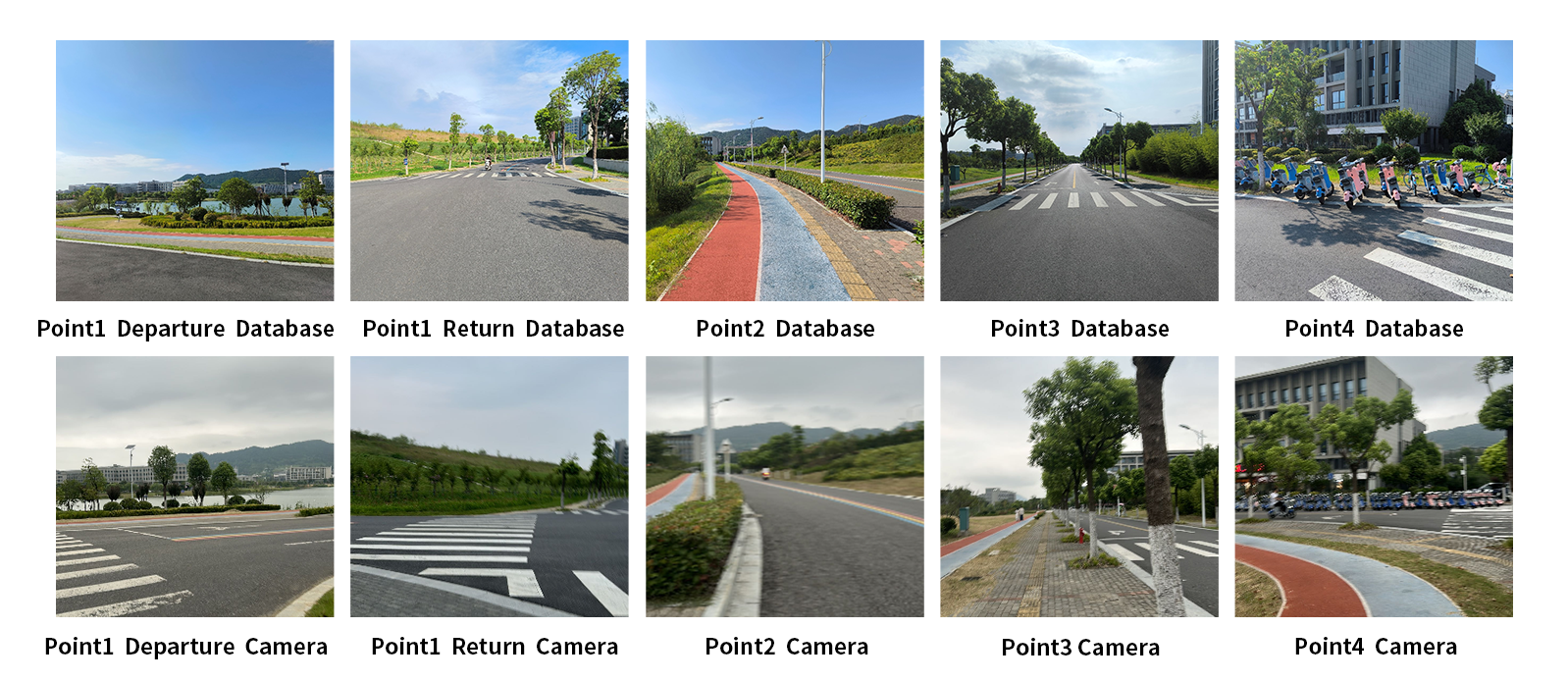}
    \caption{Example for Unrecognized Visual Beacons in Trajectory 1}\label{0827_example}
\end{figure*}

We begin by presenting the evaluation results obtained in the lakeside environment, as illustrated in Figure \ref{0827_traj}. These results include both the proposed MDR-PDR trajectory and the fused PDR-VPR trajectory. In the sparse feature environment, we set a total of 11 visual beacons, of which 7 were successfully recognized (marked as Red Points) while 4 were unrecognized (marked as Green Points), as shown in Figure \ref{0827_example}. 

\begin{figure}[t]
  \centering
   \includegraphics[width=1\linewidth]{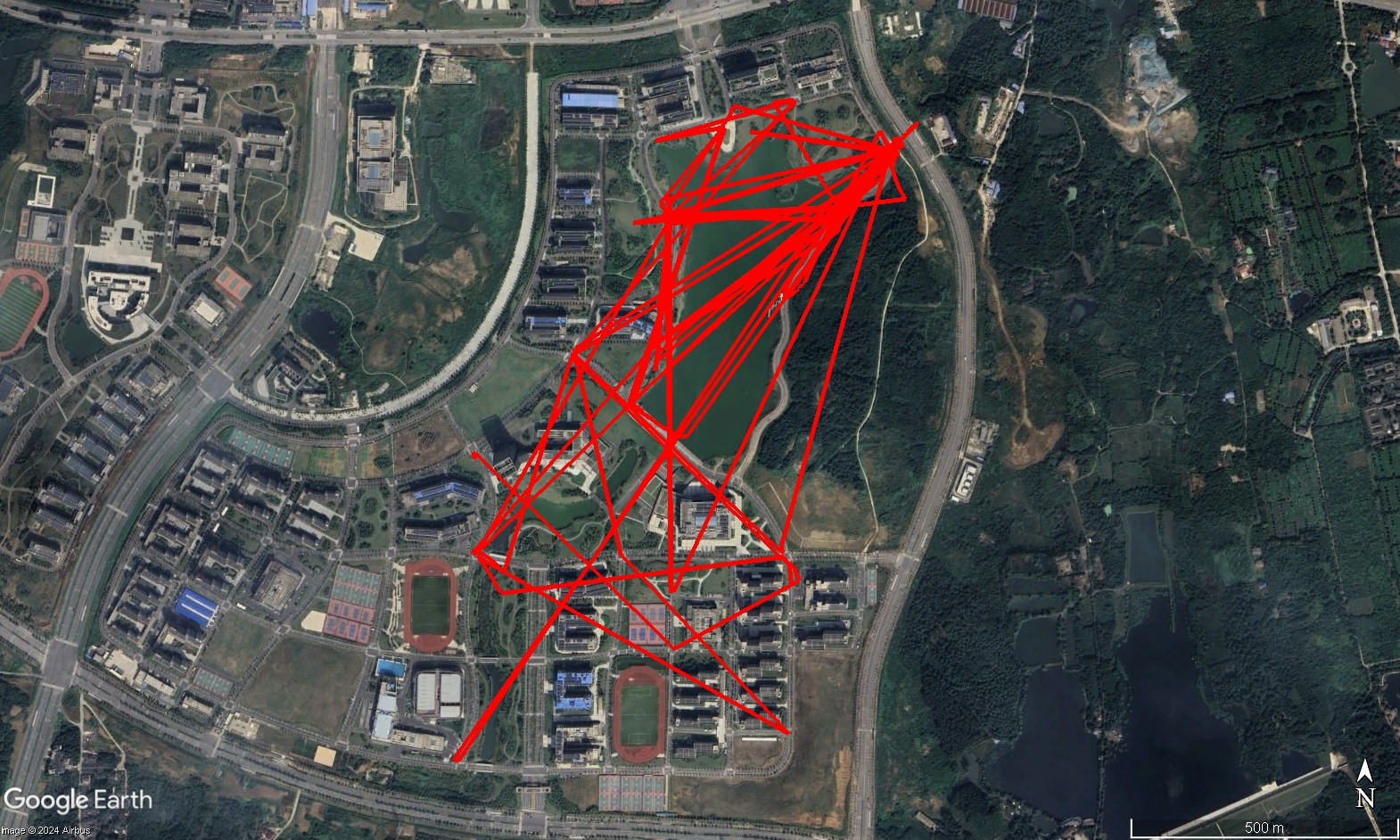}
    \caption{PDR-VPR without Gross Error Suppression for Trajectory 1}\label{0827_noThre}
\end{figure} 

We aim to clarify the factors influencing this result. As shown in Figure \ref{0827_example}, we provide a comparison of the actual landmark images corresponding to four unrecognized points. The unrecognized points may be attributed to several factors: the walking route actually passed through the visual beacons (Points 3 and 4), as well as a lack of distinctive features (Points 1 and 2). Additionally, to demonstrate the effectiveness and necessity of our gross error suppression method, we separately present the trajectory results corrected by the top-1 image from the proposed VPR model. As shown in Figure \ref{0827_noThre}, without PDR prior information as a constraint, using image similarity alone as the recognition criterion leads to completely unusable localization results. This is due to the limited Recall@1 accuracy of the lightweight VPR network, particularly when the training and testing datasets are from completely different sources (our method was trained solely on the Pitts30k-train dataset). The results in Table \ref{tab: Tokyo247} further support this phenomenon. Once a recognition error occurs, its impact on the subsequent localization results is catastrophic.

\begin{figure}[t]
  \centering
   \includegraphics[width=1.1\linewidth]{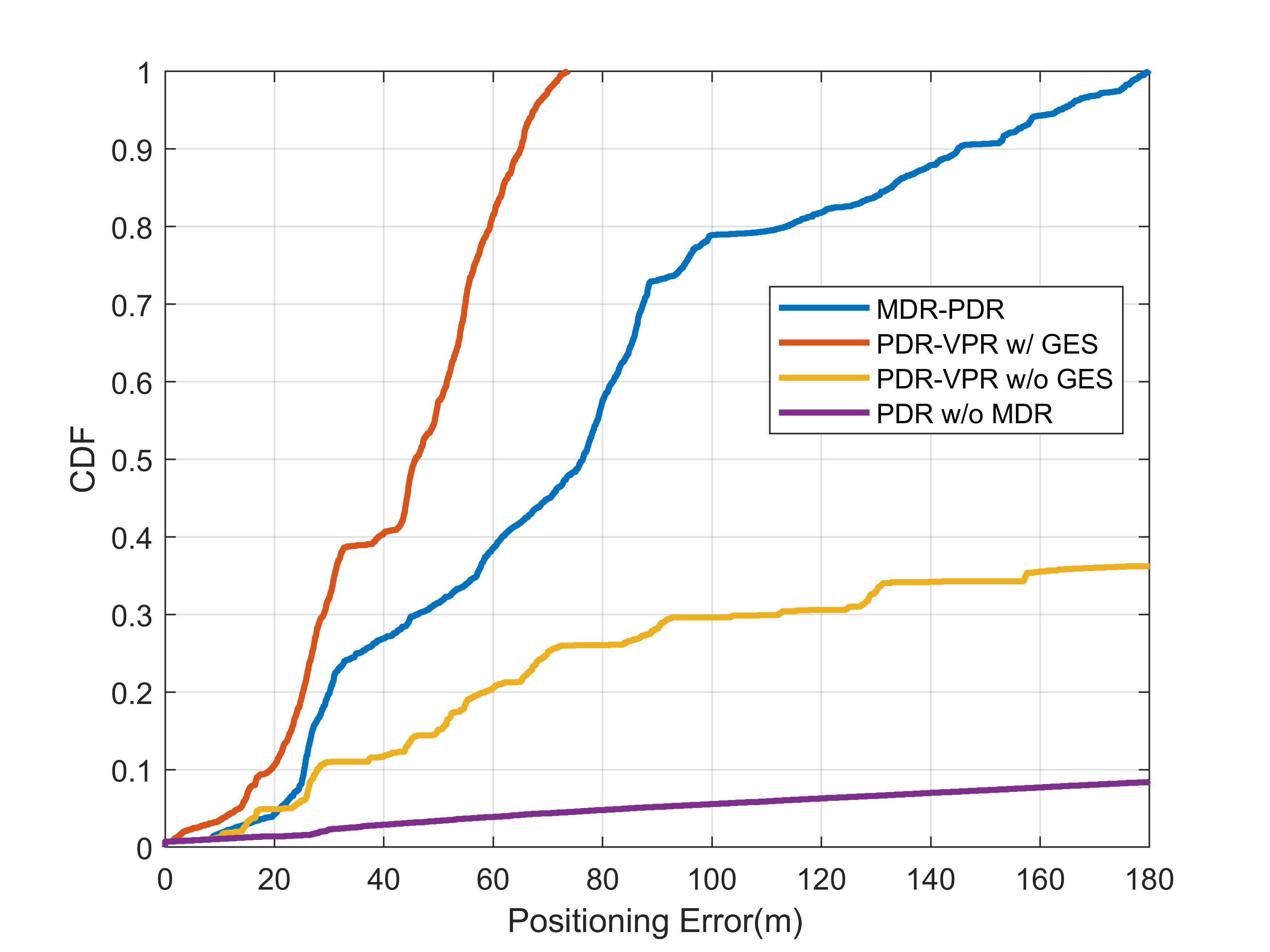}
    \caption{CDF Comparison for Trajectory 1. PDR-VPR w/ GES refers to the proposed visual-inertial localization algorithm; PDR-VPR w/o GES is the visual-inertial method without gross error suppression; MDR-PDR represents the proposed dead-reckoning method; PDR w/o MDR is the PDR algorithm without the MDR module.}\label{0827_cdf}
\end{figure}

With the aid of visual beacons, the fused trajectory aligns more closely with the ground truth compared to the MDR-PDR trajectory. This improvement is particularly evident in Figure \ref{0827_cdf}, which displays the cumulative distribution function (CDF) curves for the fused trajectory (PDR-VPR w/ GES), fused trajectory without gross error suppression (PDR-VPR w/o GES), the MDR-PDR trajectory (MDR-PDR), and the PDR trajectory without MDR (PDR w/o MDR), highlighting the enhancements made.

\begin{figure}[!t]
  \centering
   \includegraphics[width=1.1\linewidth]{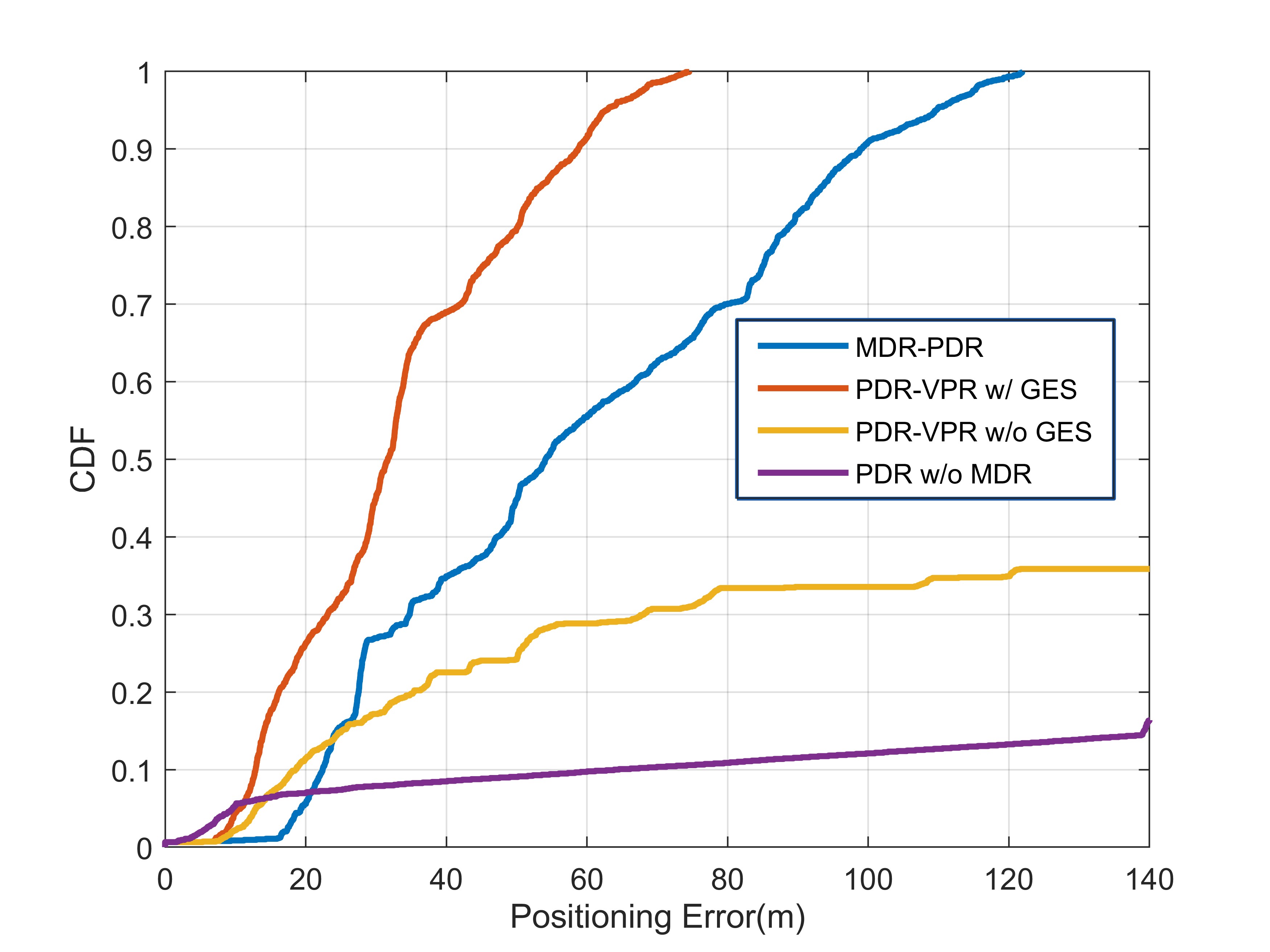}
    \caption{CDF Comparison for Trajectory 2. PDR-VPR w/ GES refers to the proposed visual-inertial localization algorithm; PDR-VPR w/o GES is the visual-inertial method without gross error suppression; MDR-PDR represents the proposed dead-reckoning method; PDR w/o MDR is the PDR algorithm without the MDR module.}\label{1112_cdf}
\end{figure}

The proposed visual-inertial algorithm demonstrated improved performance in environments with dense features. As illustrated in Figure \ref{1112_traj}, out of 12 visual beacons, 11 were successfully recognized (indicated by the Red Points), while only one beacon failed to be recognized (shown as the Green Point), resulting in a recognition accuracy of 91.7\%. The failed point is shown in the zoomed image in Figure \ref{1112_traj}, where the problematic beacon is situated among numerous repetitive trees, which obscure the nearby architectural features. Dense features are beneficial for more successful visual beacon recognition, and will contribute to enhancements in the fused trajectory. Figure \ref{1112_cdf} presents the CDF curves of the trajectories within the dense feature environment, showcasing these improvements.







\section{Conclusion}\label{}

This paper aims to provide stable and continuous positioning in large-scale urban environments by combining opportunistic visual landmark recognition and inertial pedestrian navigation, targeting potential applications on future mobile devices such as AR. To address this, we propose a lightweight MSGC-NetVLAD network for visual place recognition, which utilizes multi-scale convolutions and group convolutions to ensure a lightweight yet high-accuracy VPR. Additionally, a PDR method based on magnetic disturbance rejection is introduced to ensure positioning continuity. To address gross error issues in VPR, we propose using the natural divergence of the PDR system as a confidence threshold to constrain VPR observation updates. All modules are integrated using a Kalman filter framework. Extensive experiments demonstrate that our proposed MSGC-NetVLAD network achieves competitive accuracy. In comparison to the lightweight MobileNetV3-based VPR method, our proposed VPR solution enhances Recall@1 by at least 3\% across two public datasets while reducing the number of parameters by 63.37\%. Additionally, it achieves a performance level comparable to the larger VGG16-based method on the Pitts30k-test dataset, all with approximately 1.11 M parameters. Our overall PDR-VPR fusion algorithm improves the positioning accuracy of the proposed MDR-PDR by up to 46.86\% at a confidence level of 75\%.

However, the current work is limited to the algorithmic level, and we hope to integrate the proposed algorithm into mobile devices in the future. We believe this research will make a significant contribution to the application of AR technology and the development of smart cities.

\section*{Acknowledgements}
The authors would like to express their sincere gratitude to the editor and reviewers for their helpful comments throughout the review process, which significantly enhanced the quality of this paper. We also extend our thanks to Ms. Zhou Fengyi for her contributions to the design of the illustration graphs included in this work. Furthermore, I would like to thank Dr. Feng Chunying for conducting part of the experiments.


\bibliographystyle{cas-model2-names}

\bibliography{main}



\end{document}